\documentclass[preprint,review,12pt]{elsarticle}

\usepackage{amssymb}
\usepackage{multirow}%
\usepackage{graphicx}%
\usepackage{subcaption}
\usepackage{amsmath,amssymb,amsfonts}%
\usepackage{amsthm}%
\usepackage{mathrsfs}%
\usepackage[title]{appendix}%
\usepackage{xcolor}%
\usepackage{textcomp}%
\usepackage{manyfoot}%
\usepackage{booktabs}%
\usepackage{algorithm}%
\usepackage{algorithmicx}%
\usepackage{algpseudocode}%
\usepackage{listings}%
\usepackage{makecell}%
\usepackage{booktabs}
\usepackage[section]{placeins}
\usepackage{multido}
\usepackage[colorlinks=true, urlcolor=blue]{hyperref}

\journal{Pattern Recognition}

\newcommand{\rept}[2]{\multido{}{#1}{#2}}
\newcommand{\rightdash}[1][6]{\;\rept{#1}{\textcolor{lightgray}{\textemdash}}}
\newcommand{\leftdash}[1][6]{\rept{#1}{\textcolor{lightgray}{\textemdash}}\;\;}
\newlength\longest

\begin{document}

\begin{frontmatter}
\


\title{Document Image Cleaning using Budget-Aware Black-Box Approximation}


\author[inst1]{Ganesh Tata\corref{cor1}}
\ead{gtata@ualberta.ca}

\affiliation[inst1]{organization={Department of Computing Science, University of Alberta},
            addressline={116 St \& 85 Ave}, 
            city={Edmonton},
            postcode={T6G 2R3}, 
            state={Alberta},
            country={Canada }}

\author[inst1]{Katyani Singh}
\ead{katyani@ualberta.ca}
\author[inst2]{Eric Van Oeveren}
\ead{eric.vanoeveren@intuit.com}
\author[inst1]{Nilanjan Ray}
\ead{nray@ualberta.ca}

\cortext[cor1]{Corresponding Author}

\affiliation[inst2]{organization={Intuit Inc.},
            addressline={2700 Coast Ave}, 
            city={Mountain View},
            postcode={94043}, 
            state={California},
            country={USA}}


\begin{abstract}
Recent work has shown that by approximating the behaviour of a non-differentiable black-box function using a neural network, the black-box can be integrated into a differentiable training pipeline for end-to-end training. This methodology is termed ``differentiable bypass,'' and a successful application of this method involves training a document preprocessor to improve the performance of a black-box OCR engine. However, a good approximation of an OCR engine requires querying it for all samples throughout the training process, which can be computationally and financially expensive. Several zeroth-order optimization (ZO) algorithms have been proposed in black-box attack literature to find adversarial examples for a black-box model by computing its gradient in a query-efficient manner. However, the query complexity and convergence rate of such algorithms makes them infeasible for our problem. In this work, we propose two sample selection algorithms to train an OCR preprocessor with less than 10\% of the original system's OCR engine queries, resulting in more than 60\% reduction of the total training time without significant loss of accuracy. We also show an improvement of 4\% in the word-level accuracy of a commercial OCR engine with only 2.5\% of the total queries and a 32x reduction in monetary cost. Further, we propose a simple ranking technique to prune 30\% of the document images from the training dataset without affecting the system's performance.
\end{abstract}


\begin{highlights}
\item Black-box OCR engines can perform better with a trained document image preprocessor. 
\item Training such a preprocessor requires several queries to the OCR engine. 
\item We propose sample selection methods to reduce the OCR engine queries drastically. 
\item A simple ranking scheme is used to train the system with a smaller subset of data. 
 
\end{highlights}


\end{frontmatter}



\section{Introduction}\label{sec1}

In some machine learning systems, a non-differentiable black-box function is approximated using a differentiable surrogate model to facilitate end-to-end training using gradient-based methods \citep{jacovi2019neural}\citep{diffBypassThesis}\citep{nguyen2020end}\citep{randika2021unknown}. The effectiveness of this ``differentiable bypass'' approach can be particularly evidenced through the improvement in performance achieved by training a preprocessor for a black-box Optical Character Recognition (OCR) engine \citep{randika2021unknown}. A neural network called the "approximator" is trained to approximate the behaviour of the black-box OCR engine. However, training a good black-box approximator requires several queries of the OCR engine, which can be computationally expensive for open-source engines like Tesseract\footnote{https://github.com/tesseract-ocr/tesseract} and EasyOCR\footnote{https://github.com/JaidedAI/EasyOCR}, or incur a high financial cost for proprietary OCR APIs like the Google Cloud Vision API\footnote{https://cloud.google.com/vision/docs/ocr}. In this regard, our work focuses on reducing the number of queries of an OCR engine when training a surrogate model for efficient training of the OCR preprocessor.


    

For OCR, several commercial and open-source solutions have been made available. Commercial OCR systems are usually trained on many different types of documents since they are used for various client use cases, making them powerful OCR engines. However, fine-tuning commercial OCR APIs is not straightforward, while fine-tuning open-source OCR engines requires a good understanding of their re-training process, which can be cumbersome. The differentiable bypass \citep{randika2021unknown} approach significantly improves the performance of OCR engines like Tesseract and EasyOCR without fine-tuning the engine itself. While this approach boosts OCR performance, there is an associated tradeoff regarding queries made to the OCR engine, which can incur high financial/computational costs for training, especially when APIs like Google Vision API are used. The high costs associated with training such a preprocessor underscore the need for improving the query efficiency of the differentiable bypass system to achieve good text recognition performance at a fraction of the system's total OCR queries.


Zeroth-order optimization (ZO) techniques have been used to estimate the gradient of a function by querying it at different points without using the function's first order derivative \citep{chen2019adamm}\citep{gh2013zosgd} \citep{signsgd}. ZO methods have been particularly effective for query-efficient black-box attacks \citep{skipjumpattack}\citep{signopt}\citep{ilyas2018black}. However, such methods require \textit{at least two evaluations} of the function for each sample to estimate the gradient of the function. The convergence rate of many such methods also becomes higher as the dimensionality of the parameter space increases. Hence, ZO methods are \textit{not suitable} for approximating the OCR with fewer queries. The score function approximator approach (SFE) \citep{randika2021unknown} trains the preprocessor for the OCR by computing the gradient of the black-box OCR using the REINFORCE algorithm \citep{reinforce1992}. However, even with 10 queries of the OCR for each sample in every epoch, the system's performance does not match that of the differentiable bypass approach. Hence it is clear that the SFE method would perform worse in a low-query regime.

\begin{figure}[ht]
    \centering
    \includegraphics[width=0.95\columnwidth]{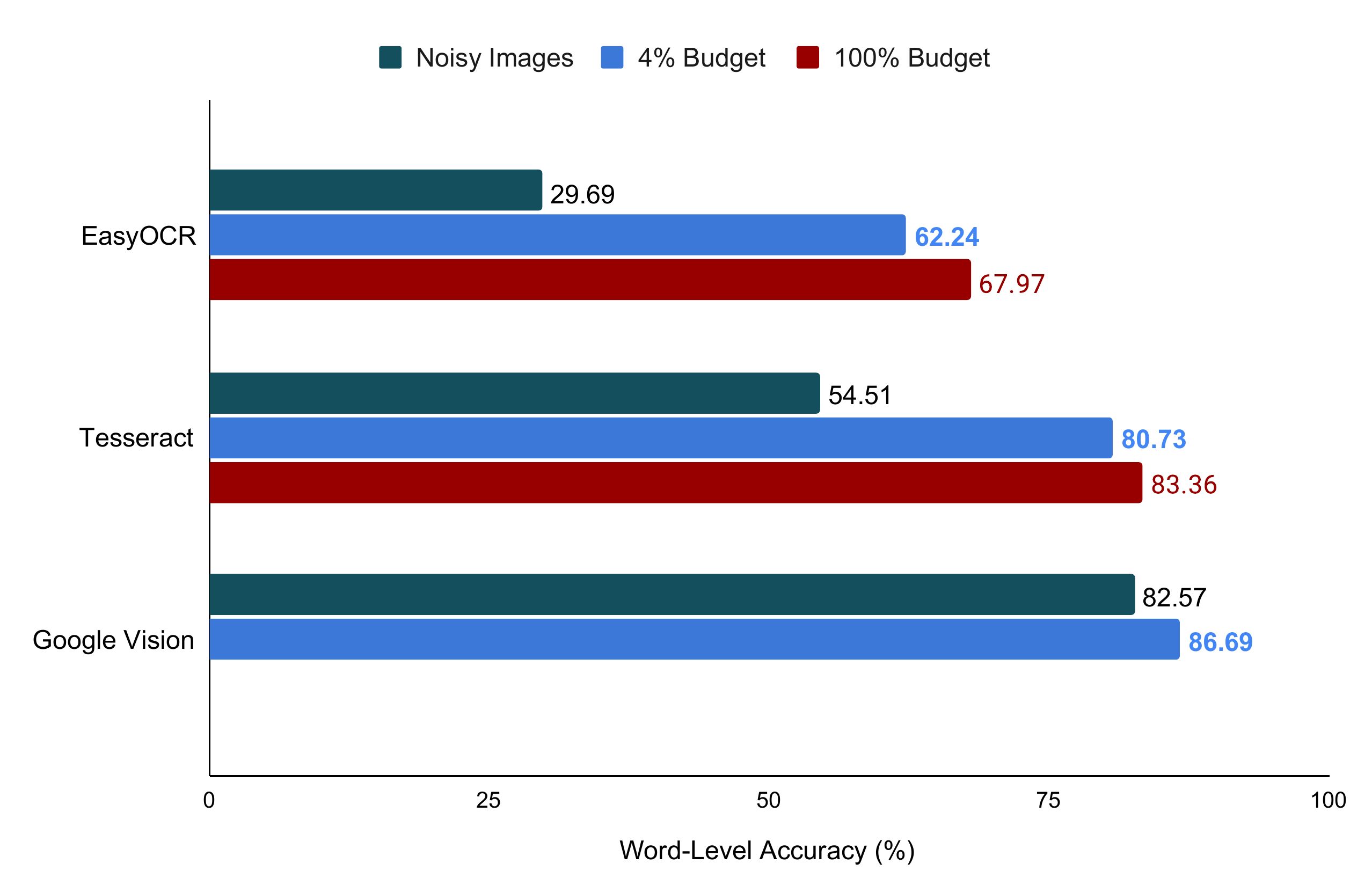}
    \caption{Test set performance of OCR engines on noisy and preprocessed images for the POS dataset with 4\% and 100\% of the total OCR queries using UniformCER selection. Even though the Google Vision API performs well for the noisy images, our methodology further improves its performance in a cost-efficient manner. We do not report results for Google Vision API with 100\% budget since the monetary cost of running that experiment is too high (Table \ref{table:gvision}).} 
    \label{fig:low_query_perf}
\end{figure}

Recently, data-efficient training of neural networks has become a key area of research. While recent advances in deep learning have achieved state-of-the-art performance in several domains, they have also led to huge computational costs, increased carbon footprint, high financial costs, and increased training time \citep{sharir2020cost} \citep{strubell2019energy}. Training models with smaller subsets of data can yield faster training cycles and a quicker turnaround time for hyperparameter tuning. Some recent scholarship has focused on adaptive subset selection \citep{gradmatchkillam} to choose a smaller number of samples for training neural networks without degrading performance. These methods include core-set selection \citep{mirza20coresets}, gradient-based scoring \citep{paul2021el2n}, and loss-based prioritization \citep{selectBackprop}. Inspired by these methods, we propose two simple selection algorithms, UniformCER and TopKCER, which select a small subset of samples in each training mini-batch for querying the OCR engine and for subsequently updating the parameters of the approximator. Both UniformCER and TopKCER select samples by utilizing the approximator's Character Error Rate (CER), which is based on Levenshtein Distance \citep{levenshtein1966binary}, to measure the hardness of each sample. Additionally, these selection algorithms add no extra computational overhead to the system. As shown in Fig. \ref{fig:low_query_perf}, sample selection using UniformCER with a very low query budget (4\%) can improve the text recognition performance for different OCR engines. We also propose a simple technique to prune document images before training the system to reduce the OCR engine queries further. Pruning is performed by ranking document images using the OCR engine's CER and removing a proportion of low ranking images such that the system can be trained with the pruned dataset without a significant change in the text recognition performance.

Thus, in this work we propose two selection methods, UniformCER and TopKCER, to query the OCR engine for a smaller subset of samples without significant reduction in OCR performance and with an overall query budget of less than 10\%.  We demonstrate that UniformCER and TopKCER methods outperform random sampling for two low query budgets. We also show that increasing the query budget beyond 10\% leads to a larger training time for the system and marginal improvement in  OCR performance. Finally, we use a simple ranking technique to prune 30\% of the receipt images from the training dataset without significant reduction in test accuracy for the trained
preprocessor. The source code used for conducting this research work is available at \href{https://github.com/tataganesh/Query-Efficient-Approx-to-improve-OCR}{github.com/tataganesh/Query-Efficient-Approx-to-improve-OCR}.

\section{Related Work}
\subsection{Differentiable Bypass for Black-Box Integration}
Differentiable bypass has been used in various applications to integrate a black-box into a differentiable training pipeline. EDPCNN \citep{nguyen2020end} trains a CNN for left ventricle segmentation by approximating the behaviour of a non-differentiable Dynamic Programming module using a neural network. Similarly,  EstiNet \citep{jacovi2019neural} is a general framework for training a differentiable estimator as a proxy for a black-box function to facilitate the composition of different black-box functions and trainable modules. Further, differentiable proxies have been used to optimize the hyperparameters of black-box image signal processing (ISPs) units and circumvent manual configuration when deploying imaging systems for different applications \citep{Tseng2019isp}. 

Differentiable bypass has also been effective in enhancing the performance of black-box OCR engines. Preprocessing images for OCR is essential to improve its text recognition performance. Commonly used preprocessing methods involve image binarization \citep{deepotsu}\citep{sauvolaadaptive}\citep{chen2008double}, independent component analysis \citep{ICA_cleaning}, and deep learning-based super-resolution \citep{deepSR} \citep{peng2020building}. However, these methods do not preprocess the image for the specific OCR engine being used. Tuning them to work well for different OCR engines is also cumbersome. Thus, a  customized pre-processor \citep{randika2021unknown} is trained by approximating the gradient of a black-box OCR engine using a neural network, which significantly improves OCR performance.

\subsection{Accelerated Neural Network Training}
Several algorithms have been proposed to prioritize samples for accelerating the training of neural networks. Importance sampling techniques have been used to select \citep{fernandez_IS}  \citep{katharopoulos_IS} important samples for accelerating neural network training and reducing computational resource utilization. Some methods use curriculum learning \citep{bengiocurriculum} to choose good subsets of data for training. Minimax curriculum learning selects hard and diverse samples using submodular maximization, and the diversity of chosen subsets is adjusted based on different stages of training \citep{zhou2018minimax}. Dynamic Instance Hardness (DIH) is a measure for quantifying the hardness of each sample during the training process to reduce the training time of neural networks using curriculum learning \citep{zhou2020curriculum}. They propose the per-sample loss as a  metric to compute the DIH for each training sample. Similarly, Selective-Backprop \citep{selectBackprop} performs an additional forward pass to obtain the loss for each sample in a mini-batch and selects samples with higher loss to compute the gradient. The Coresets for Accelerating Incremental Gradient descent (CRAIG) \citep{mirza20coresets} algorithm selects a weighted subset of training samples using submodular maximization of the facility location function \citep{krause2014submodular} to approximately estimate the full gradient of the original training set with respect to a loss function.  Selection-via-proxy \citep{coleman2019selection} uses a low-capacity proxy model to perform coreset selection from a given training dataset in a computationally efficient manner for accelerating the training of a larger capacity model. In an active learning setting, uncertainty sampling techniques like Max Entropy are used to select the most uncertain samples for annotation \citep{burr_AL}. Further, filtered active submodular selection (FASS) \citep{weisubmodular} is a batch active learning algorithm that selects diverse samples from the set of most uncertain samples for a given model. 

\section{Background}
We briefly describe the setup mentioned in \citep{randika2021unknown}. A preprocessor $g$ with parameters $\theta$ is trained to perform transformations on the input document image $x$. During training, each image is first passed through $g$ to obtain a preprocessed image $g(x)$. If the document image has more than one word, the words are cropped from $g(x)$. Then, Gaussian noise $\epsilon$ is added to the word images to ``jitter'' them and pass them through an OCR engine and its differentiable approximator $f$ with parameters $\phi$. The jitter prevents overfitting in the approximator by providing some exploration in the input space. The output from the OCR engine is used as labels to match the output of $f$ and update its parameters. On the other hand, to train $g$, the ground truth labels are used, the parameters of $f$, i.e., $\phi$, are frozen, and $f$ is used as a differentiable proxy for the OCR. Initially, $f$ is trained separately with the output of the OCR engine to obtain good initial weights for the system, thereby avoiding the cold-start problem. This alternating training scheme was proposed as Estinet \citep{jacovi2019neural}. Approximating the black-box OCR engine using the differentiable approximator enables end-to-end training of the system without requiring intermediate labels to train the preprocessor. The training system has been illustrated in Fig. \ref{fig:overall_pipline}.




\section{Methods}
\subsection{Overview}
\begin{figure}[h]
        \includegraphics[width=\columnwidth]{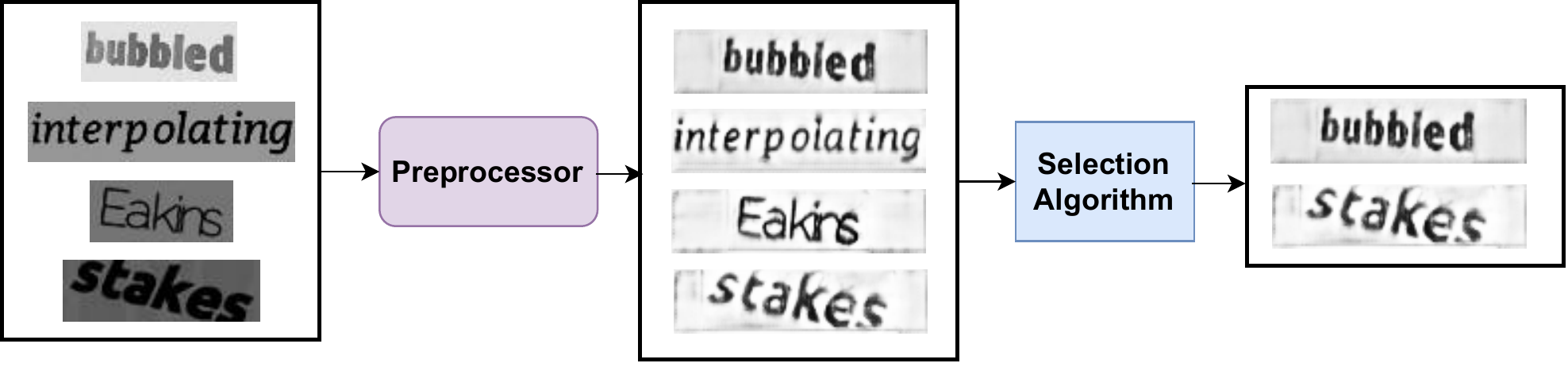}
        \vspace{0.5mm}
        \caption{\textbf{Sample selection}. Samples are chosen using a selection algorithm after images are passed through the preprocessor.}
        \label{fig:sample_selection}
\end{figure}

Algorithm \ref{algorithm:1} shows how the selection algorithms reduce OCR queries. First, a small subset of preprocessed samples is selected from each mini-batch before passing through the approximator and OCR. Subsequently, Gaussian noise is added to the selection samples, which are passed through the OCR for computing the loss function to update $\phi$. Finally, the preprocessor is updated using the entire minibatch, and this is possible because of differentiable bypass. To the best of our knowledge, except for EstiNet \citep{jacovi2019neural}, which is also a neural network-based approximator, no ZO method would allow us to update the preprocessor by skipping OCR calls. In this section, we elaborate on the sample selection mechanisms. Fig. \ref{fig:sample_selection} provides an overview of sample selection. 

\begin{figure}[h]
    \includegraphics[width=\columnwidth]{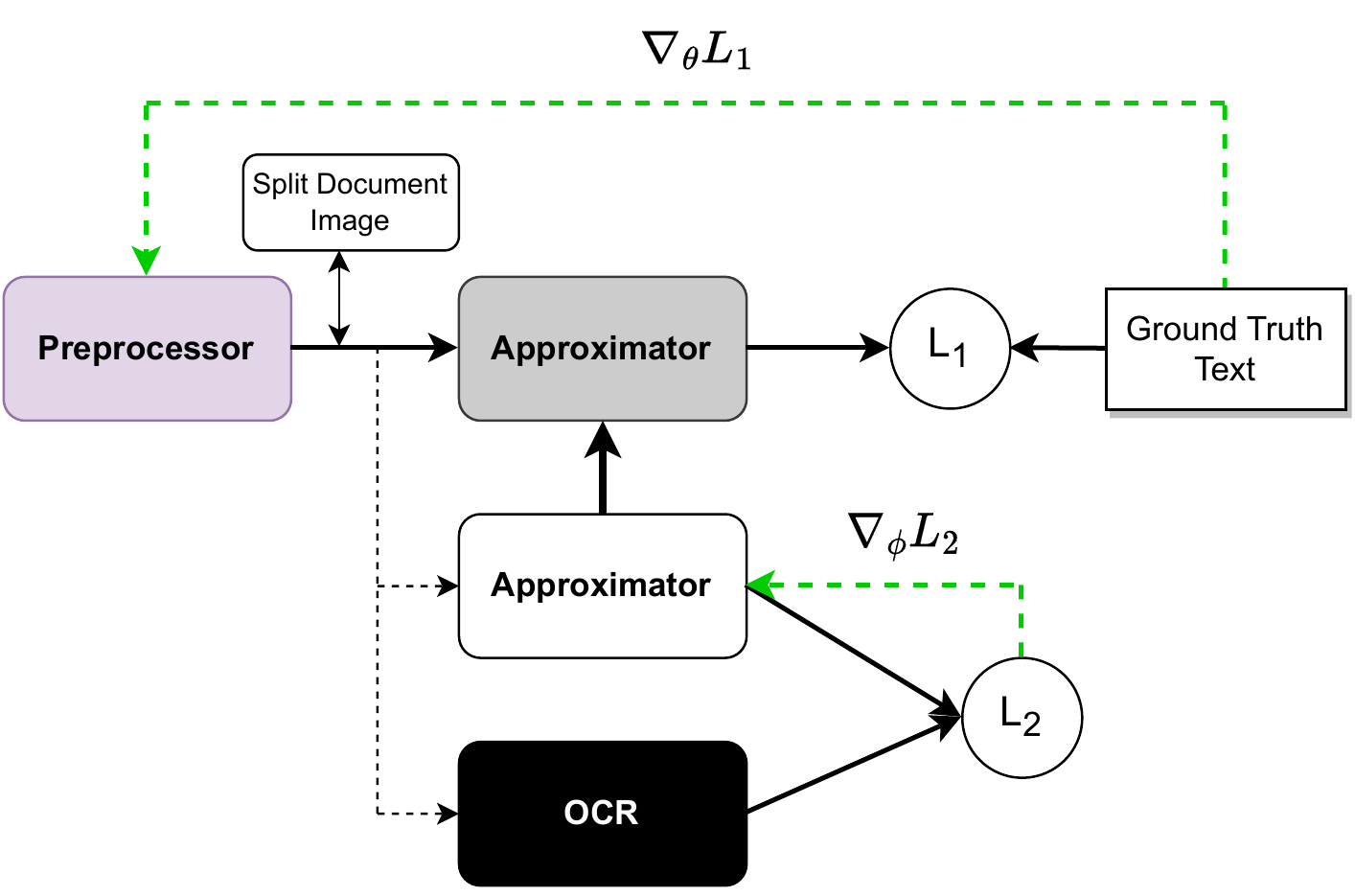}
    \vspace{0.5mm}
   
    \caption{\textbf{Training pipeline}. An overview of the training pipeline \cite{randika2021unknown}. The broken green lines indicate backpropagation while the broken black lines depict only forward propagation to illustrate the training of the preprocessor using differentiable bypass. Receipt and document images are split into text strips after preprocessing.}
    \label{fig:overall_pipline}
\end{figure}

\subsection{Selecting Samples for OCR}
\label{sec:sample_selection}
Obtaining OCR labels for each text strip can be computationally expensive (for open-source engines) or costs money (commercial software/APIs). In our setup, the preprocessed word images are passed as a mini-batch $B$ to $f$ and the OCR engine. The goal of subset selection is to choose a representative subset of size $k$ from each mini-batch such that  $k \ll |B|$  and obtain the OCR labels for them to train $f$. Our approach for sample selection involves choosing a representative subset \citep{zhou2018minimax}\citep{weisubmodular}\citep{kirsch2019batchbald} based on a measurable property of each sample. In this regard, we propose two techniques for selecting samples to obtain OCR labels - UniformCER and TopKCER.

\begin{algorithm}
\caption{Efficient OCR approximation for document image cleaning using sample selection}
\label{algorithm:1}
\begin{algorithmic}
\State \textbf{Input}: \textit{$X_{train}$}, \textit{$Y_{train}$}, \textit{k}, \textit{$\eta_1$}, \textit{$\eta_2$}, \textit{$\sigma$}
\For{$x_{batch}, y_{batch} \in \{\textit{$X_{train}$}, \textit{$Y_{train}$}\}$}
\State $g_{batch} = preprocessor(x_{batch})$
\State $subset = select(g_{batch}, k)$
\For{$x \in  subset$}
\State Sample $\epsilon \sim \mathcal{N}(0, \sigma)$ 
\State $L$ += CTC($f$($x$ + $\epsilon$), \textit{OCR}($x$ + $\epsilon$)) 
    

\EndFor
\State $\phi = \phi - \eta_1 \nabla_{\phi} L$
\State $\theta = \theta - \eta_2 \nabla_{\theta} L(f(x_{batch}), y_{batch})$
\EndFor

\end{algorithmic}
\end{algorithm}

\subsubsection{UniformCER}

\label{section:uniformCER}
Character Error Rate (CER) is a character-level metric used for evaluating OCR systems. We assume that CER, with respect to the ground truth label, quantifies the \textit{hardness} of each sample. The CER for a sample is calculated as shown in Eqn. \ref{eq:1},  where $n$ is the number of characters in the ground truth word, $s$ is the number of substitutions, $i$ is the number of insertions, and $d$ is the number of deletions. 
\begin{equation} \label{eq:1}
    \text{CER} = \dfrac{(s + i + d)}{n}
\end{equation}
As shown in Algorithm \ref{algorithm:1}, the parameters of preprocessor $g$ are updated by calculating the loss function between $f$ and the OCR for \textit{all} samples in mini-batch $B$, which enables the computation of CER value for each sample in $B$ with respect to the ground truth labels. These CER values are recorded in each training epoch. The CERs stored in the previous epoch for each sample in $B$ are used for sample selection to query the OCR. We obtain the minimum and maximum CER value for each mini-batch, denoted by $cer_{min}$ and $cer_{max}$, respectively. Then, $k$ values are sampled from $\text{Uniform}(cer_{min}, cer_{max})$. Finally, we determine the sample whose CER is closest to each of the $k$ selected points. These $k$ samples are passed through the OCR and the  CRNN to compute the  CTC loss function and update the weights of the CRNN. Algorithm \ref{algorithm:2} shows the mathematical details of UniformCER.


\begin{algorithm}
\caption{UniformCER Selection Algorithm}
\label{algorithm:2}
\begin{algorithmic}

\State \textbf{Input}: $x_{batch}$,  $cers_{batch}$, $k$
\State $cer_{max}$ = max($cers_{batch}$)
\State $cer_{min}$ = min($cers_{batch}$)
\State $c_1$, $c_2$ ... $c_k$ $\sim$ $U(cer_{min}, cer_{max})$
\State $idx$ = \{\}
\For{$c_i \in c_1, ... c_k$}
\State $j$ = $\underset{j \notin idx}
{\mathrm{argmin}}\, (|c_i - (x_{batch})_j|)$
\State $idx$.insert($j$)
\EndFor
\State $subset = x_{batch}[idx]$

\end{algorithmic}
\end{algorithm}

\subsubsection{TopKCER}
\label{section:topkcer}
Selective Backprop and Variance Reduction Importance Sampling (VR) perform loss-based sample prioritization to select a subset of samples for neural-network training acceleration. These techniques assume that samples with a higher loss are highly informative. Since CER can also be treated as a measure of informativeness because it measures the hardness of text strips, we can select samples with the highest CERs computed during training of the system (as shown in \ref{section:uniformCER}) to query the OCR engine.

\subsection{Pruning Document Images}

\begin{figure}[t]
        \includegraphics[width=\columnwidth]{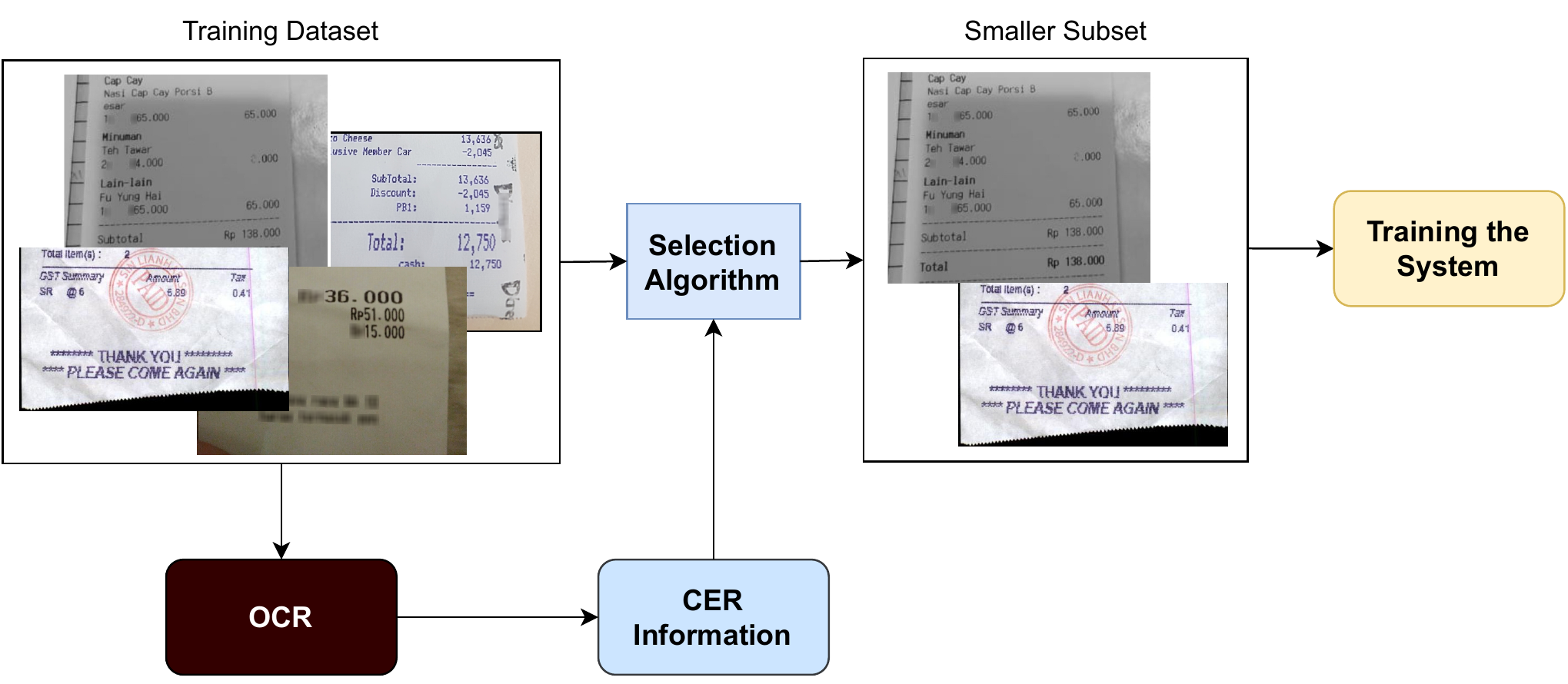}
        \vspace{0.5mm}
        \caption{Samples selection for the POS dataset  using a selection algorithm before training the preprocessor using differentiable bypass.}
        \label{fig:coreset_selection}
\end{figure}
We propose a simple technique to prune the document image dataset and train the preprocessor using a smaller subset of documents. In our document cleaning setup, the approximator is pre-trained with the output of the OCR engine to avoid the cold start problem. The OCR engine is queried once for all text strips/word images in the dataset to obtain labels for pre-training. We use these OCR string labels to compute the CER of each text strip with respect to the Ground Truth labels. This CER quantifies the OCR engine's capability to predict the correct characters in each text strip. A text strip with a high CER can be considered a \textit{hard} sample for the OCR engine, while a text strip with a low CER can be considered an \textit{easy} sample. Then, we compute the \textit{mean CER} for each document image using the CER of the text strips present in the image. The mean CER gives us a ranking of the images based on the OCR engine's CER. Images are selected by choosing the top-k samples based on mean CER. Performing sample selection from the dataset enables us to remove document images before training the system. The techniques proposed in section \ref{sec:sample_selection} can be combined with our pruning algorithm to reduce queries to the OCR engine further. Fig. \ref{fig:coreset_selection} provides an overview of the data pruning setup. 
\section{Experiment Setup}
\subsection{Datasets}
 All experiments are performed on the ``POS dataset'' and the VGG dataset, curated as shown in \citep{randika2021unknown}, with Tesseract and EasyOCR as the black-box OCR engines. The POS dataset is a combination of three POS (Point-of-Sale receipt) datasets - Findit fraud detection dataset \citep{findit}, ICDAR SROIE competition dataset \citep{icdarSOIR} and CORD dataset \citep{park2019cord}. Image patches are extracted from the receipt images and then resized such that the images in the dataset have a maximum height of $400$ pixels and a width of 500 pixels. With roughly 90k word patches, the POS dataset has 3676 training images, 424 validation images, and 417 test images. The VGG dataset consists of 60k word images randomly sampled from the VGG synthetic word dataset \citep{jaderberg2014synthetic}. The dataset is split into 50k training images, 5k validation images, and 5k test images. 
 \subsection{Training Details}
Similar to \citet{randika2021unknown}, a UNet architecture \citep{ronneberger2015u} is considered for the preprocessor, while a CRNN architecture \citep{shi2016end} is used for the approximator. The system is trained using the Adam optimizer \citep{kingma2014adam} for 50 epochs, with the learning rates for the preprocessor and approximator being $5 \times 10^{-5}$ and $10^{-4}$ respectively. A weight decay of $5 \times 10^{-4}$ is used for both the approximator and preprocessor when the POS dataset is used for training. The CRNN model is pre-trained with the OCR for 50 epochs to avoid the cold start problem \citep{randika2021unknown}.  For noise jitter, $\sigma$ is randomly sampled from $0, 0.01, 0.02, 0.03, 0.04$, and $0.05$. Most of these hyperparameter values are picked up from the best values used to train the original system \citep{randika2021unknown}. The full receipt images are first passed through the preprocessor when training the system with the POS dataset. Then, each preprocessed image is split into word images to pass them through the approximator and to query the OCR engine since our CRNN model can only work with word images as inputs. Hence, for receipt images, the batch size for the preprocessor is 1, and the batch size for the approximator is the number of text strips in the preprocessed document image. For the VGG dataset, a batch size of 64 is used for the UNet and the CRNN. The CTC loss function \citep{graves_ctc} is used to update both networks in the system. The mean squared error (MSE) between the preprocessed image and a white image (image with all ones) is used as a secondary loss function when updating the preprocessor to "whiten" the image \citep{randika2021unknown}. The parameter $\beta$ controls the effect of the MSE loss function on the total loss. We use $\beta=1$ for our experiments \citep{randika2021unknown}. Thus, the preprocessor loss function is given by
 \begin{equation}
     L_{prep}(x, y_{gt}) = \text{CTC}(f(g(x), y_{gt}) + \beta * \text{MSE}(g(x), J_{m \times n})
 \end{equation}
 Since our aim is to reduce the number of queries to the OCR engine, we do not perform extensive hyperparameter sweeps for all settings. In most cases, the hyperparameters from \citep{randika2021unknown} were sufficient for the system to achieve good performance. However, for EasyOCR and POS dataset, we obtained sub-par performance with the default hyperparameters. Hence, we tuned the learning rate of UNet and CRNN for 4\% query budget and TopKCER selection using the open-source library Optuna \citep{akiba2019optuna}. Using the best-performing hyperparameters (learning rate of UNet and CRNN are $5 \times 10^{-4}$ and $1.5 \times 10^{-5}$ respectively), we performed the rest of the experiments. The results for EasyOCR on POS dataset are reported with these hyperparameters. 
 
 The preprocessor's weights are updated by calculating the loss with respect to the ground truth labels for all samples in a mini-batch (Fig. \ref{fig:overall_pipline}). To train the approximator, a small subset of samples is selected from the preprocessed images (or the text strips extracted from a preprocessed image). Sample selection is performed before querying the OCR engine such that the number of queries in each epoch is $n\%$ of the queries in the original system per epoch. In the original setup, Gaussian noise is added twice for each sample, resulting in two queries per sample in each epoch. Hence, if $n=10$, we select 20\% of the total text strips in each minibatch to query the OCR and update CRNN's weights (10\% of (2  $|\text{text strips}|$) = 20\% of $|\text{text strips}|$). In our work, we consider $n \in \{4, 8\}$ to demonstrate the efficacy of our approach on low query budgets. There are no additional hyperparameters associated with the proposed selection methods. We consider random sampling as a baseline method for our experiments. For UniformCER and TopKCER, we compute the CER of Tesseract for all text strips before training the system and use it for performing selection in the first epoch. 
 
\begin{figure}[]
    \begin{subfigure}[t]{0.5\textwidth}
         \centering
         \includegraphics[width=\textwidth]{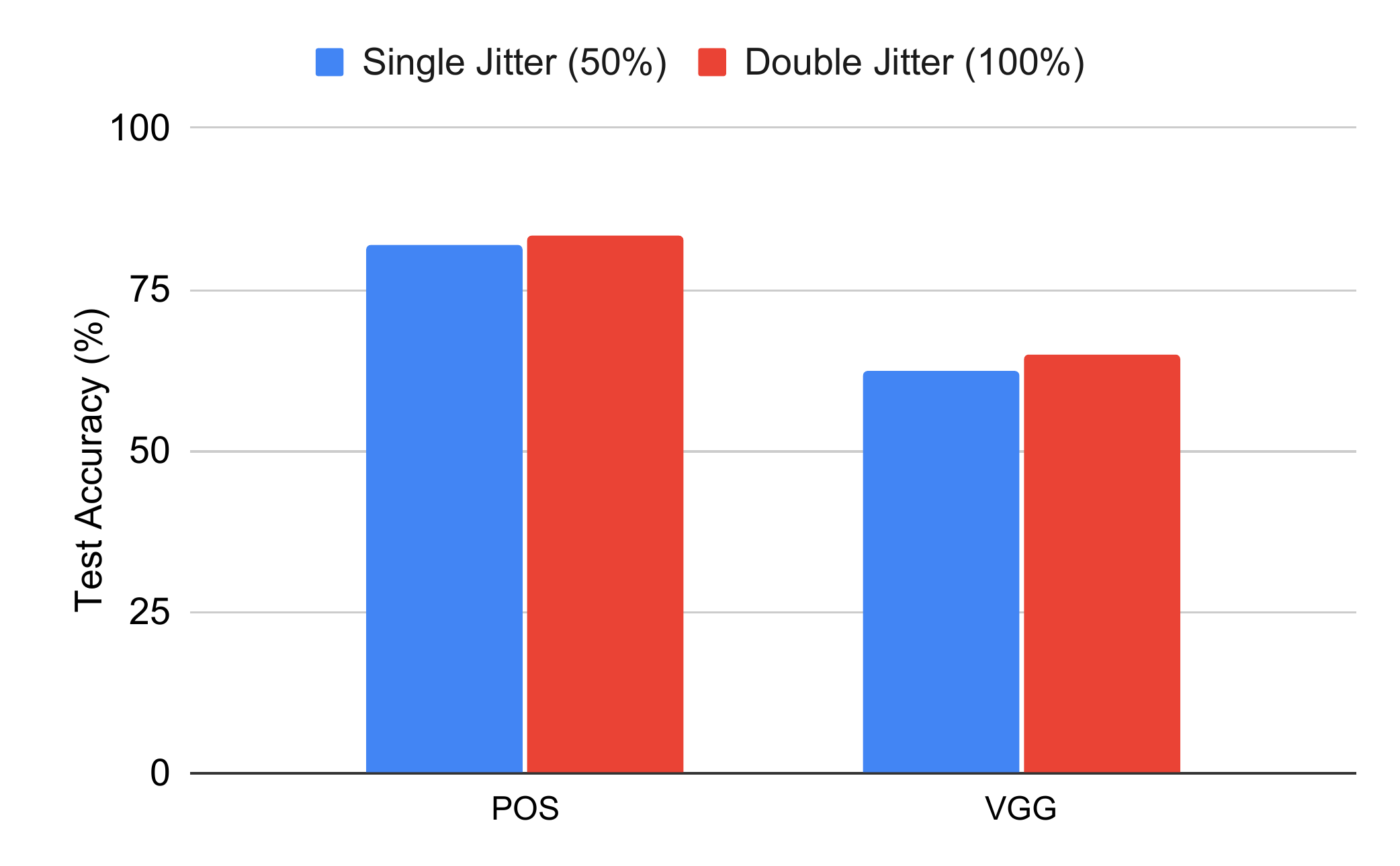}
         \caption{Tesseract}
         \label{subfig:test-full-half}
     \end{subfigure}%
    \begin{subfigure}[t]{0.5\textwidth}
         \centering
         \includegraphics[width=\textwidth]{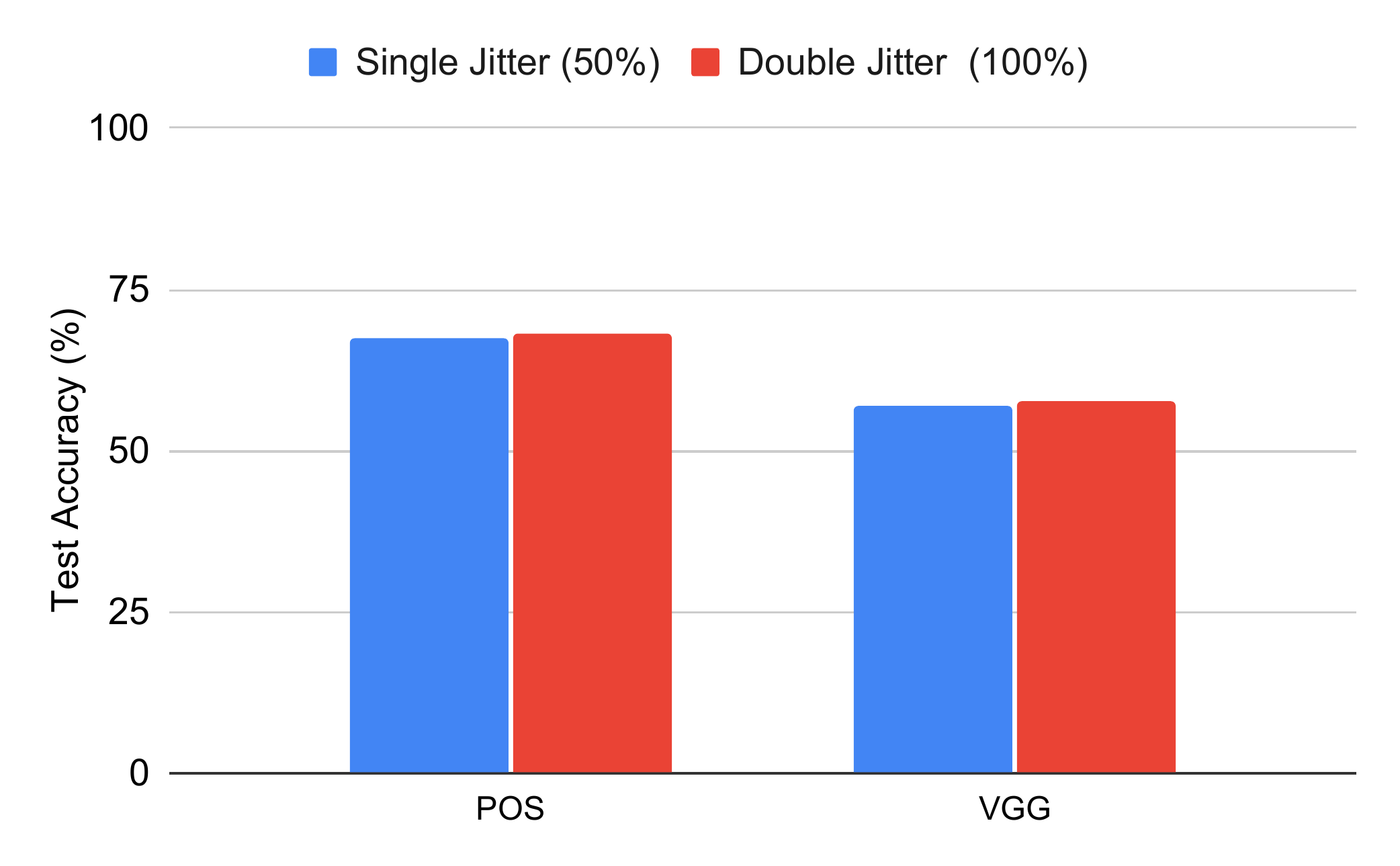}
         \caption{EasyOCR}
         \label{subfig:easyocr-full-half}
     \end{subfigure}
    \caption{Test set word-level accuracy (\%) of OCR engine  with noise added once (50\% budget) and twice (100\% budget) for POS and VGG datasets. }
    \label{fig:full-half-budget}
\end{figure}

 We refer to the results presented in \citep{randika2021unknown} for 100\% budget, while the results for 50\% budget are obtained by adding Gaussian noise to the samples only once. In this work, we jitter the samples only once since it allows us to reduce the number of queries by 50\% with a minimal drop in accuracy (Fig. \ref{fig:full-half-budget}). The OCR is not queried for experiments with a 0\% budget, and the weights of the pre-trained CRNN model are frozen throughout training. Our system's evaluation metric is the OCR engine's word-level accuracy with the pre-processed images. It calculates the proportion of recognized words that correctly match the ground truth text. The word accuracy is reported using the preprocessor checkpoint with the highest validation accuracy. All models were trained on an NVIDIA V100 GPU. 

\aboverulesep=0ex
\belowrulesep=0ex
\renewcommand{\arraystretch}{1}

\textbf{Google Vision API.} Since it costs money to query the Google Vision API\footnote{https://cloud.google.com/vision/pricing}, we only perform experiments with the POS dataset, UniformCER selection, and 2.5\% budget. 2.5\% is the minimum budget possible in this setup since we observed that the OCR needs to be queried \textit{for at least one sample} in each minibatch to achieve good performance on low query budgets. Further, querying the OCR for all samples in the validation set in each epoch incurs a high cost, so we randomly sample 50 images ($\sim$800 text strips) from the validation set and use the word-level accuracy on this image subset to choose the best model checkpoint. Finally, we train this system for 41 epochs.
\section{Results}\label{sec2}
\begin{figure}[H]
\centering
\begin{subfigure}{\textwidth}
    
        \centering
        \includegraphics[width=0.85\textwidth]{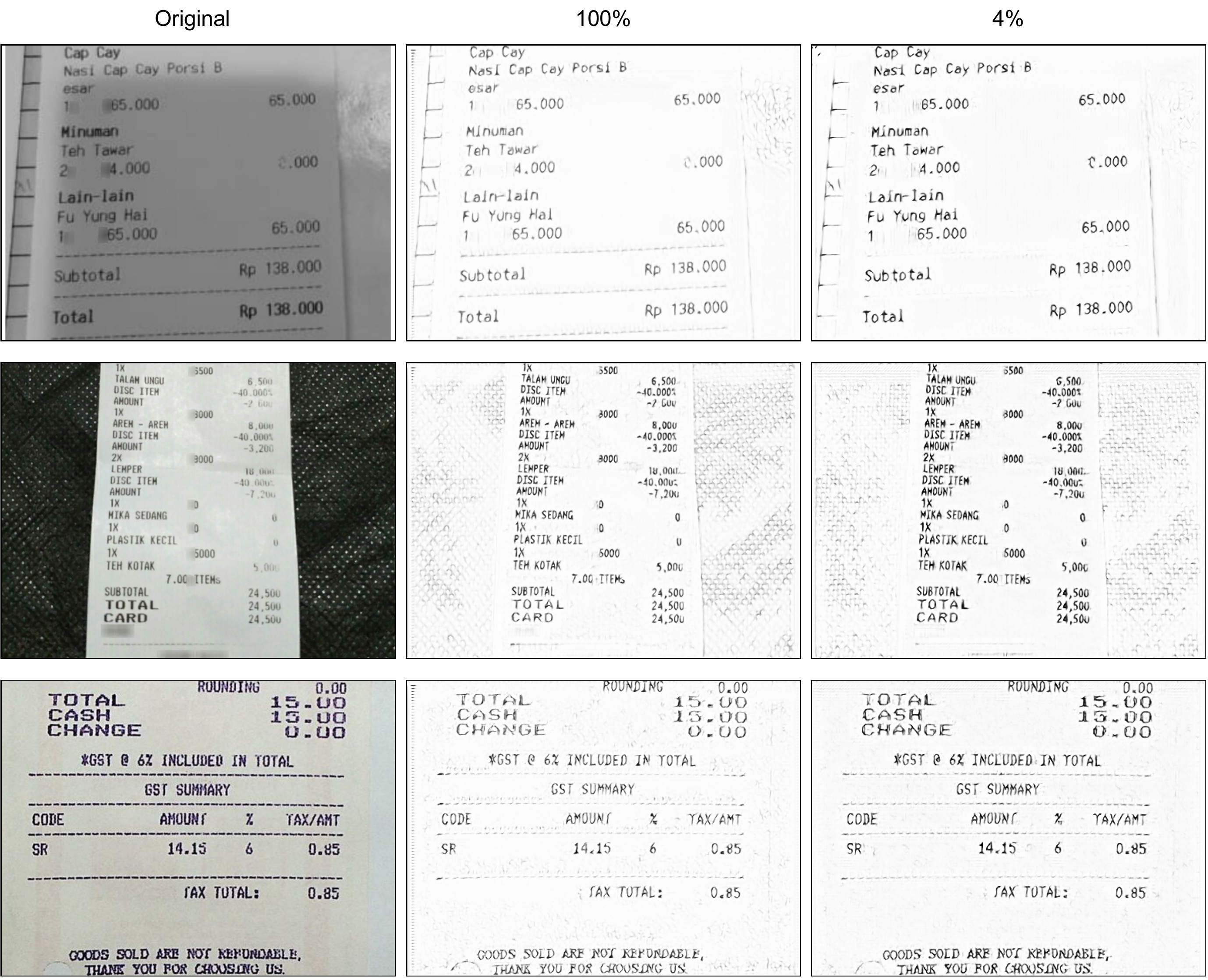}
        \label{pos_collage}
        \caption{Preprocessed images for POS dataset.}

\end{subfigure}\\[4ex]

\begin{subfigure}[b]{0.85\textwidth}
        \centering
        \includegraphics[width=0.85\textwidth]{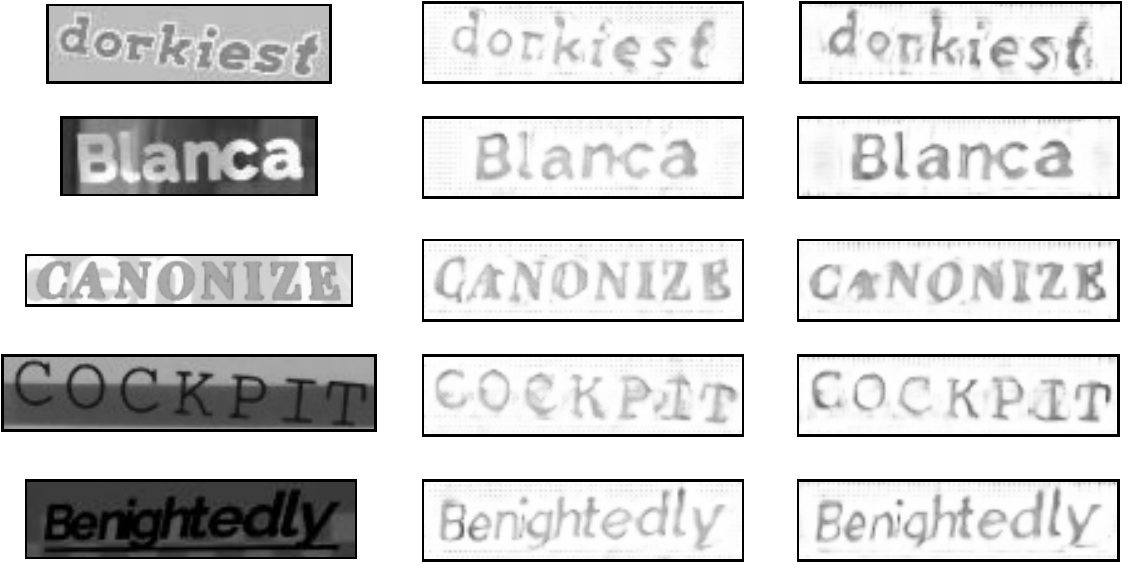}
        \label{fig:vgg_collage}
        \caption{Preprocessed images for VGG dataset.}

\end{subfigure}
\caption{ Preprocessor output for different test samples in the  a) POS dataset and b) VGG dataset. The preprocessors were trained with Tesseract OCR engine and UniformCER selection was used for 4
\% budget.  \textbf{Column 1:} Original Images. \textbf{Column 2:} Original system - 100\% query budget. \textbf{Column 3:} 4\% query budget.}
\label{fig:prep_images_all}
\end{figure}

Tables \ref{table:tess_results} and \ref{table:easyocr_results} depict the word-level accuracy of different selection algorithms on Tesseract and EasyOCR, respectively, across two query budgets for both datasets. Fig. \ref{fig:prep_images_all} shows the preprocessed images obtained after training the preprocessor with different query budgets.

\textbf{Importance of querying the OCR}.  For most settings, it is evident that with a budget of only 4\%, the performance of all selection algorithms is better than the performance of the system with no OCR engine queries (0\% query budget). Further, selection using random sampling alone leads to improvement in OCR performance with less than 10\% of the query budget, which shows that random sampling is a strong baseline.  These results also indicate that updating the approximator with at least a few queries to the OCR engine is essential to train a better preprocessor. 

\begin{table}[ht]
\small
\centering
\resizebox{0.8\columnwidth}{!}{%
\begin{tabular}{@{}ccccc@{}}
\toprule
\multirow{2}{*}{\textbf{Dataset}} & \multirow{2}{*}{\textbf{Budget}} & \multicolumn{3}{c}{\textbf{Selection Method}}                                     \\ \cmidrule(l){3-5} 
                     &                      & Random                 & UniformCER             & TopKCER                         \\ \midrule[1pt]
\multirow{5}{*}{POS} & 4\% & 78.36 & 80.73 & \textbf{81.6} 
                                          \\ \cmidrule(l){2-5}
                     & 8\% & 79.2  & 81.34 & \textbf{81.44}
                                          \\ \cmidrule(l){2-5} 
                     & 0\% & \multicolumn{3}{c}{\leftdash 75.23 \rightdash}                                                  \\  
                     & 100\%                 & \multicolumn{3}{c}{\leftdash 83.36 \rightdash}                                   
                      
            \\ \midrule[1pt]
\multirow{5}{*}{VGG} & 4\% & 62.04 & 62.86 & \textbf{63.5}   
                                     \\  \cmidrule(l){2-5} 
                     & 8\%           & 62.16 & \textbf{63.86} & 63.56 
                                    \\  \cmidrule(l){2-5} 

                     & 0\%                  & \multicolumn{3}{c}{\leftdash 45.60 \rightdash}                                                         \\
                     & 100\%                 & \multicolumn{3}{c}{\leftdash 64.94 \rightdash}

                     \\ \bottomrule
\end{tabular}%
}
\caption{Word-level accuracy (\%) with Tesseract for different selection methods. No selection methods were used for 0\% and 100\% budgets. }
\label{table:tess_results}

\end{table}
\textbf{Importance of querying the OCR}.  For most settings, it is evident that with a budget of only 4\%, the performance of all selection algorithms is better than the performance of the system with no OCR engine queries (0\% query budget). Further, selection using random sampling alone leads to improvement in OCR performance with less than 10\% of the query budget, which shows that random sampling is a strong baseline.  These results also indicate that updating the approximator with at least a few queries to the OCR engine is essential to train a better preprocessor. 
\begin{table}[ht]
\small
\centering
\resizebox{0.8\columnwidth}{!}{%
\begin{tabular}{@{}ccccc@{}}
\toprule
                                   &                                   & \multicolumn{3}{c}{\textbf{Selection Method}}                                                 \\ \cline{3-5} 
\multirow{-2}{*}{\textbf{Dataset}} & \multirow{-2}{*}{\textbf{Budget}} & Random                  & UniformCER                       & TopKCER                          \\ \midrule[1pt]
                                   &  4\% &  62.24 & \textbf{65.56} & 64.22 \\ \cline{2-5} 
                                   & 8\%            & 65.27 & 65.02 & \textbf{65.86}  \\ \cline{2-5} 
                                   & 0\%                               & \multicolumn{3}{c}{\leftdash 59.10 \rightdash}                                                                      \\  
 \multirow{-3}{*}{POS}             
                                   & 100\%                              & \multicolumn{3}{c}{\leftdash 67.97 \rightdash}                                                                      \\ 
                                  
                                    \midrule[1pt]
                                & 4\% &  56 &  56.12 & \textbf{56.84} \\ \cline{2-5}
       \multirow{-1}{*}{VGG}    & 8\%       & 56.62 & 57.8   & \textbf{57.82} \\   \cline{2-5}                
                                   & 0\%                               & \multicolumn{3}{c}{\leftdash 49.0 \rightdash}                                                                        \\ 
                                & 100\%  & \multicolumn{3}{c}{\leftdash 57.48 \rightdash}  \\   
                        \bottomrule
\end{tabular}%
}
\caption{Word-level accuracy (\%) with EasyOCR for different selection methods. No selection methods were used for 0\% and 100\% budgets. }
\label{table:easyocr_results}
\end{table}
\textbf{Performance of selection algorithms}. Across the board, both UniformCER and TopKCER perform better than random sampling. This is particularly evident for the 8\% budget setting with the POS dataset for both Tesseract and EasyOCR. Compared to the original system with 100\% query budget, UniformCER and TopKCER with 8\% query budget results in a  3\% or lower drop in accuracy for both OCR engines. For the VGG dataset, all selection algorithms display a minimum improvement of 7\% when compared to the performance at  0\% budget. For this dataset, all selection algorithms  have 2\% or lower drop in accuracy for both 4\% and 8\% budgets, with both CER-based algorithms performing better than random sampling. From these results, we can conclude that for lower budgets, selection using a sample measure like CER is necessary to achieve 1-2\% improvement over random sampling.\\
\textbf{Google Vision API Results.} Table \ref{table:gvision} shows the result for training a preprocessor with the Google Vision API for the POS dataset. We observe that there is 4\% increase in the word-level accuracy with just 2.5\% query budget. The increase in accuracy with a low budget shows that the performance of Google Vision API can be improved by training a preprocessor using differentiable bypass. We also observe that training the preprocessor using the original system would require \textbf{32x more cost} than training it with 2.5\% budget, which is a significant reduction in cost.
\setlength{\aboverulesep}{5pt}
\setlength{\belowrulesep}{5pt}
\begin{table}[h]
\centering
\resizebox{\columnwidth}{!}{%
\begin{tabular}{@{}ccccc@{}}
\toprule
{\textbf{Dataset}} &
  {\textbf{\parbox{3cm}{\centering Without \\ Preprocessing}}} &
  \textbf{\parbox{3cm}{\centering With \\ Preprocessing \\ Budget=2.5\%}} &
    \textbf{\parbox{3cm}{\centering Projected \\ Expense \\ Budget=100\%}} &
  \textbf{\parbox{3cm}{\centering Actual \\ Expense \\ Budget=2.5\%}}

                       \\ \midrule
POS & 82.57\% & 86.69\% & 9030 USD  & 280 USD
                      \\ \bottomrule
\end{tabular}%
}
\caption{Test set word-level accuracy (\%) of preprocessor trained with Google Vision API along with expected and actual cost of training the system. }
\label{table:gvision}
\end{table}

\section{Additional Experiments}
\subsection{Training and testing on different OCR engines}

\begin{table}[ht]
\centering
\resizebox{\columnwidth}{!}{
\begin{tabular}{ccccc} 
\hline
\textbf{Dataset} & \begin{tabular}[c]{@{}c@{}}\textbf{OCR used}\\\textbf{ for training}\end{tabular} & \begin{tabular}[c]{@{}c@{}}\textbf{OCR used}\\\textbf{ for testing}\end{tabular} & \begin{tabular}[c]{@{}c@{}}\textbf{Test Accuracy }\\\textbf{ (4\% Budget)}\end{tabular} & \begin{tabular}[c]{@{}c@{}}\textbf{Test Accuracy }\\\textbf{ (100\% Budget)}\end{tabular}  \\ 
\hline
POS              & Tesseract                                                                         & EasyOCR                                                                          & 39.06                                                                                   & 40.44                                                                                      \\ 
\hline
VGG              & Tesseract                                                                         & EasyOCR                                                                          & 43.08                                                                                   & 47.14                                                                                      \\ 
\hline
POS              & EasyOCR                                                                           & Tesseract                                                                        & 63.92                                                                                   & 60.94                                                                                      \\ 
\hline
VGG              & EasyOCR                                                                           & Tesseract                                                                        & 43.86                                                                                   & 21.64                                                                                      \\ 
\hline
POS              & Google Vision                                                                     & Tesseract                                                                        & 66.45                                                                                   & -                                                                                          \\ 
\hline
POS              & Google Vision                                                                     & EasyOCR                                                                          & 33.04                                                                                   & -                                                                                          \\ 
\hline
POS              & Tesseract                                                                         & Google Vision                                                                    & -                                                                                       & 88.23*    \\ \hline                                                                                 
\end{tabular}
}
\caption{Word-level accuracy (\%) for OCR trained and tested on different engines using best-performing preprocessors from Tables  \ref{table:tess_results},\ref{table:easyocr_results} and \ref{table:gvision}. $^{\text{\small *}}$The preprocessor was trained with 50\% query budget.}
\label{table:cross_ocr_results}
\end{table}

Table \ref{table:cross_ocr_results} shows the results for training and testing the preprocessor on different OCR engines. The results for 100\% budget have been inferred from a similar experiment conducted by  \citet{randika2021unknown}. We observe that in most cases, the difference in performance between 4\% and 100\% query budget is less than 6\% across all OCR combinations. These results demonstrate that a preprocessor trained with a low query budget performs similarly to the original system when tested on different OCR engines. Moreover, the last row shows that a preprocessor trained with Tesseract using a 50\% budget and evaluated on Google Vision API results in an accuracy improvement of 6\%. Furthermore, it is important to note that the huge increase in performance (by 20\%) observed in the fourth row of the table results from pre-training the CRNN with Tesseract instead of easyOCR for  4\% query budget on the VGG dataset. We observed that this change improves the performance of EasyOCR on a low query budget while also improving the performance on Tesseract for the VGG dataset

\begin{figure}[!htb]
    \begin{subfigure}[t]{0.5\textwidth}
         \centering
         \includegraphics[width=\textwidth]{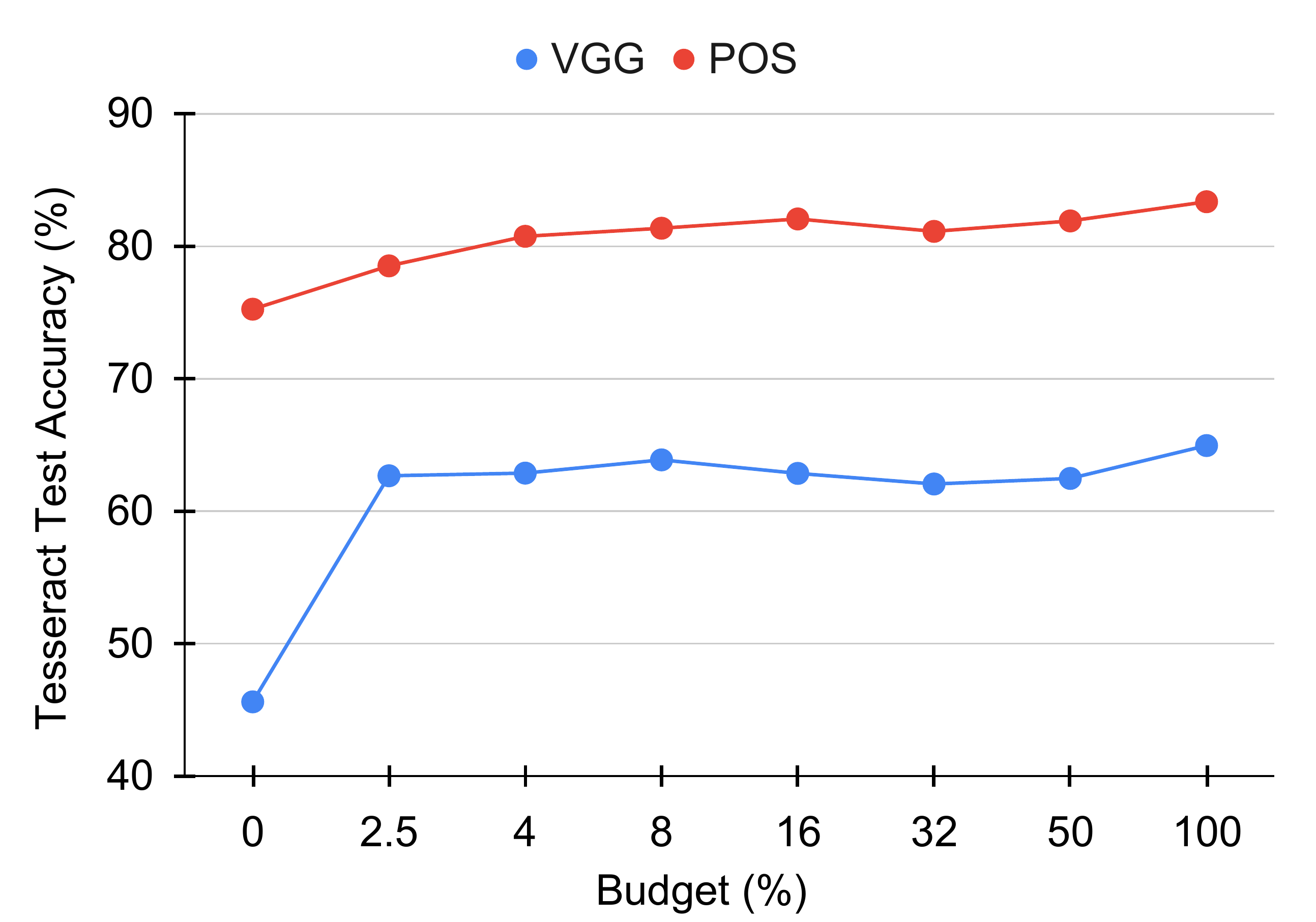}
         \caption{Test Accuracy}
         \label{subfig:test-acc-budg}
     \end{subfigure}%
    \begin{subfigure}[t]{0.5\textwidth}
         \centering
         \includegraphics[width=\textwidth]{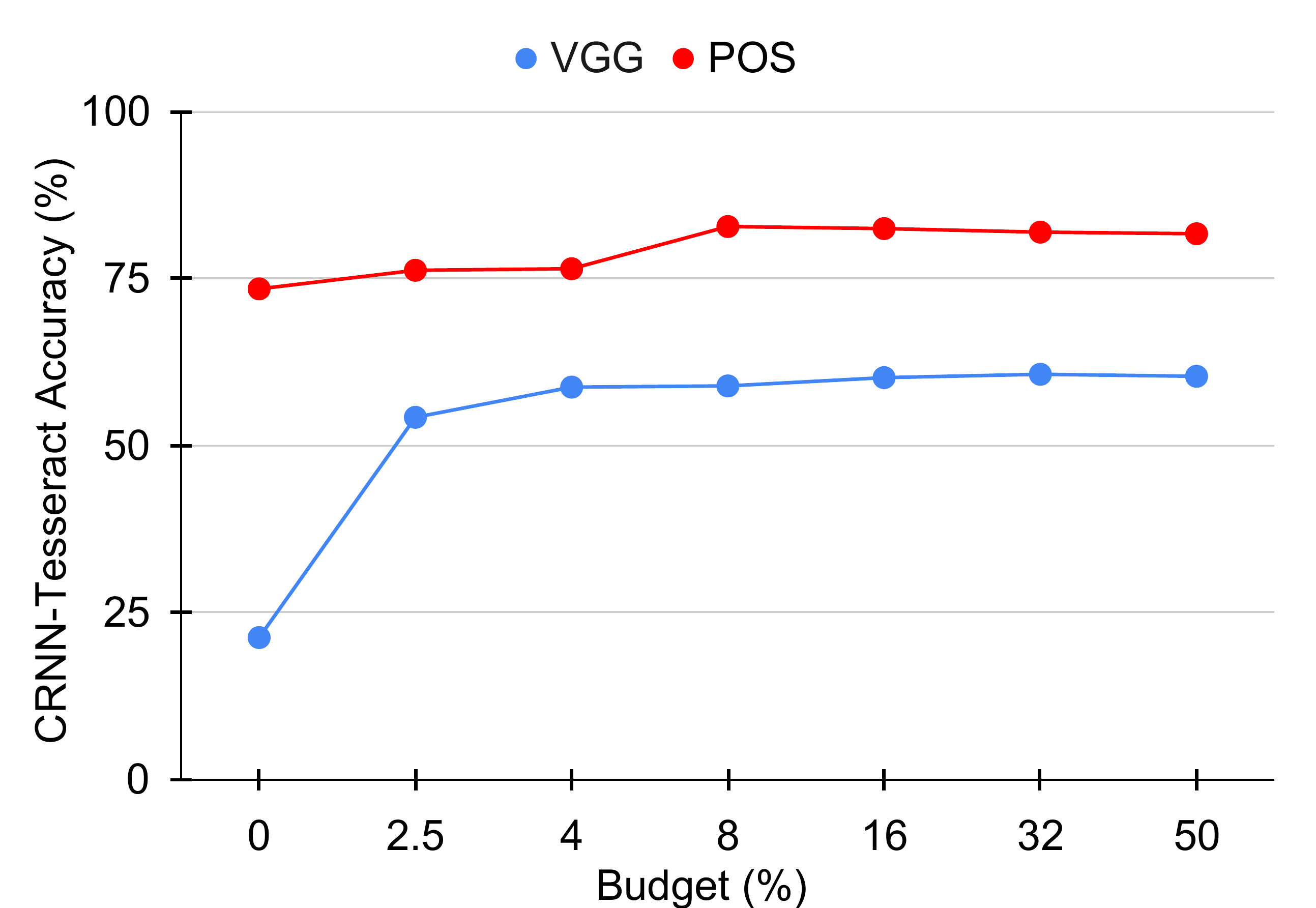}
         \caption{CRNN Accuracy}
         \label{subfig:crnn-ocr-budg}
     \end{subfigure}

    \begin{subfigure}[b]{\textwidth}
         \centering
         \includegraphics[width=0.5\textwidth]{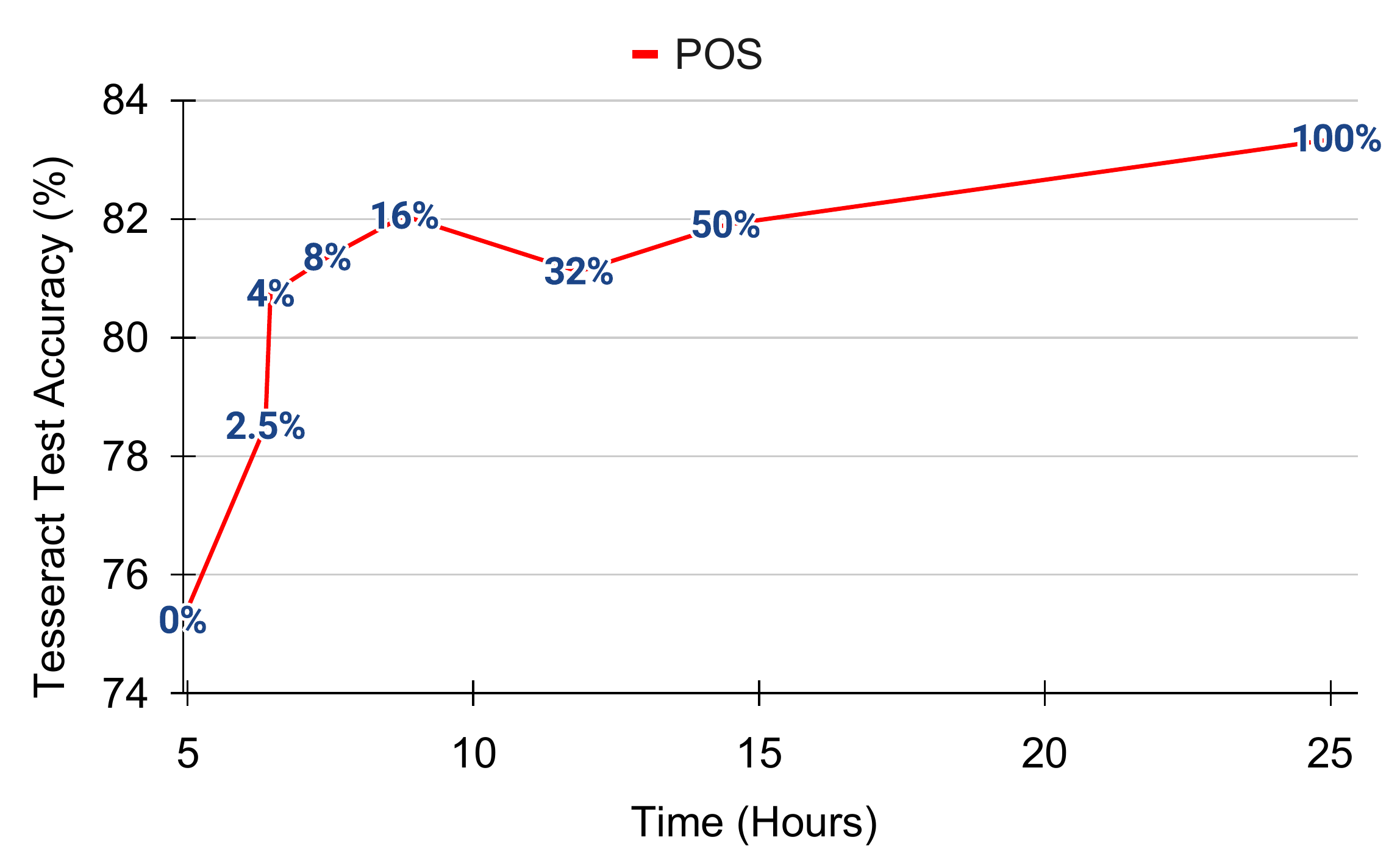}
         \caption{Training Time}
         \label{subfig:train-time-budg}
     \end{subfigure}
    \caption{(a) Test set performance, (b) CRNN accuracy with OCR predictions as ground truth and (c) system training time, with UniformCER selection and Tesseract across different query budgets. }
    \label{fig:query_budgets}
\end{figure}

\subsection{Performance across different query budgets}
Fig. \ref{subfig:test-acc-budg} depicts the performance of UniformCER with Tesseract across different query budgets on the POS dataset. We observe a clear jump in accuracy for both datasets when moving from 0\% to 2.5\% budget. However, the increase in accuracy plateaus as the budget increases beyond 4\%, indicating the diminishing return of higher query budgets on preprocessor performance. 

With the same experiment setup, we also aim to quantify the \textit{approximation strength} of CRNN with respect to the OCR engine. We determine the approximation strength by computing the accuracy of CRNN on the validation set with respect to the OCR predictions. Fig. \ref{subfig:crnn-ocr-budg} shows the \textit{CRNN-Tesseract} accuracy across different query budgets. We observe that the trend is similar to that of Fig. \ref{subfig:test-acc-budg}, which shows that the CRNN's approximation strength (with respect to Tesseract and the Patch dataset) improves with a low query budget like 2.5\%, which could potentially explain the increase in performance from 0-2.5\% in Fig.  \ref{subfig:test-acc-budg}.    

Fig. \ref{subfig:train-time-budg} shows the training time and accuracy of the system across different budgets for the POS dataset. We notice that the training time increases significantly with increase in query budget. However, these higher budgets do not lead to a significant improvement in OCR performance, indicating that we can improve the OCR performance by querying the OCR with a very low query budget.

\begin{figure}[!htb]
    \begin{subfigure}[t]{0.5\textwidth}
         \centering
         \includegraphics[width=\textwidth]{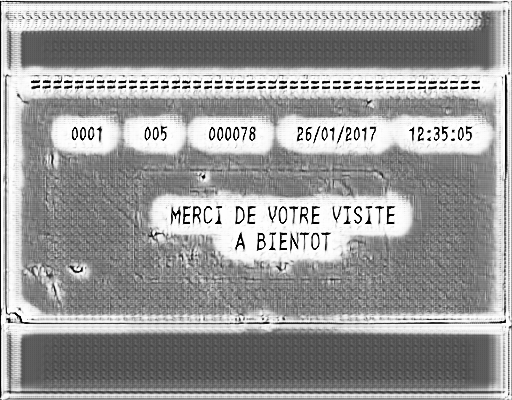}
         \caption{POS,  $\beta=0$}
         \label{subfig:pos_beta0}
     \end{subfigure}%
    \begin{subfigure}[t]{0.5\textwidth}
         \centering
         \includegraphics[width=\textwidth]{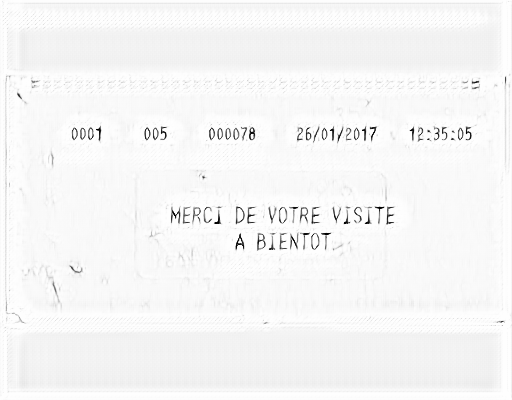}
         \caption{POS,  $\beta=1$}
         \label{subfig:pos_beta1}
     \end{subfigure} \\[2ex]

         \begin{subfigure}[t]{0.5\textwidth}
         \centering
         \includegraphics[width=\textwidth]{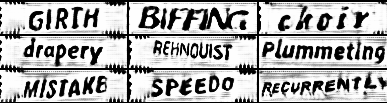}
         \caption{VGG,  $\beta=0$}
         \label{subfig:vgg_beta0}
     \end{subfigure}%
    \begin{subfigure}[t]{0.5\textwidth}
         \centering
         \includegraphics[width=\textwidth]{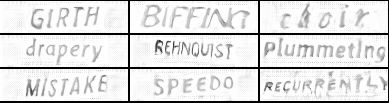}
         \caption{VGG,  $\beta=1$}
         \label{subfig:vgg_beta1}
     \end{subfigure}

    \caption{Effect of MSE loss coefficient $\beta$ on preprocessor trained with UniformCER selection, Tesseract and 8\% budget. }
    \label{fig:beta_effect}
\end{figure}

\subsection{Effect of MSE Loss}
In this section, we study the effect of the MSE loss coefficient $\beta$ when the system is trained with a low query budget. We perform experiments on both datasets with UniformCER selection, Tesseract OCR, and 8\% query budget. We vary $\beta$ from $0$ to $1$ with increments of $0.2$. For the POS dataset, we noticed that the performance does not vary significantly across different values of $\beta$. However, visual inspection of the preprocessed images produced by $\beta=0$ (Fig. \ref{subfig:pos_beta0}) and $\beta=1$ (Fig. \ref{subfig:pos_beta1}) indicates that they yield significantly different images. If the full POS receipt images are passed as input through an OCR engine (instead of text strips), the noise around the text in different parts of Fig. \ref{subfig:pos_beta0} can lead to incorrect text recognition output. Such noise is absent in Fig. \ref{subfig:pos_beta1} due to the MSE loss, making it amenable for preprocessed images to be passed through an OCR engine. A similar phenomenon is also observed for the VGG dataset, but the word-level accuracy at $\beta=0$ setting is 2\% lower than the word-level accuracy at $\beta=1$. 

\subsection{Pruning Receipt Images}
\begin{figure}
        \centering
        \includegraphics[width=0.85\columnwidth]{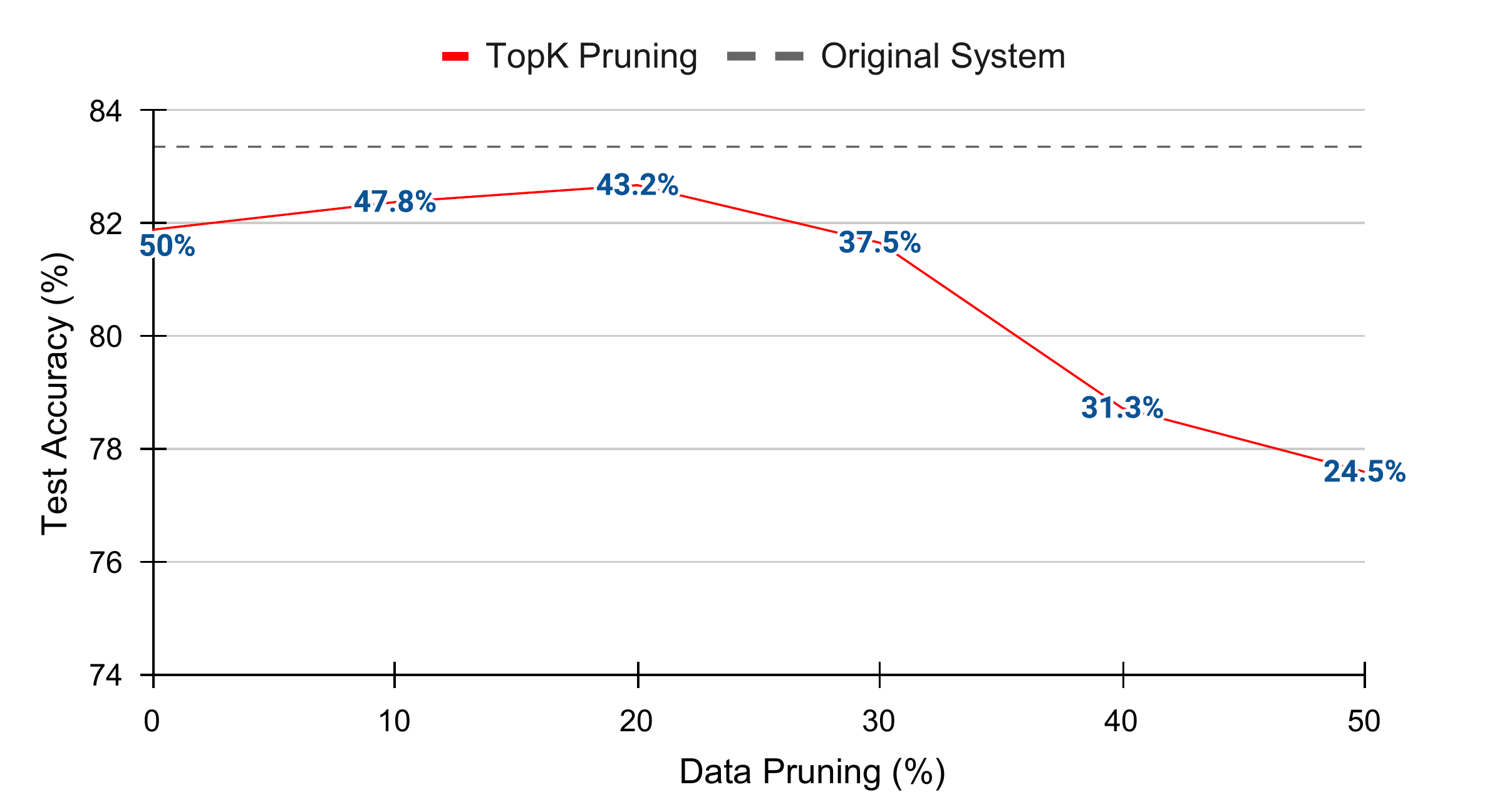}
        \vspace{0.5mm}
        \caption{Test Accuracy and Tesseract queries (\%) across different data pruning rates. The numbers in blue indicate query \% using single jitter with respect to the original system.}
        \label{fig:pruning_results}
\end{figure}

\begin{figure}
        \centering
        \includegraphics[width=0.8\columnwidth]{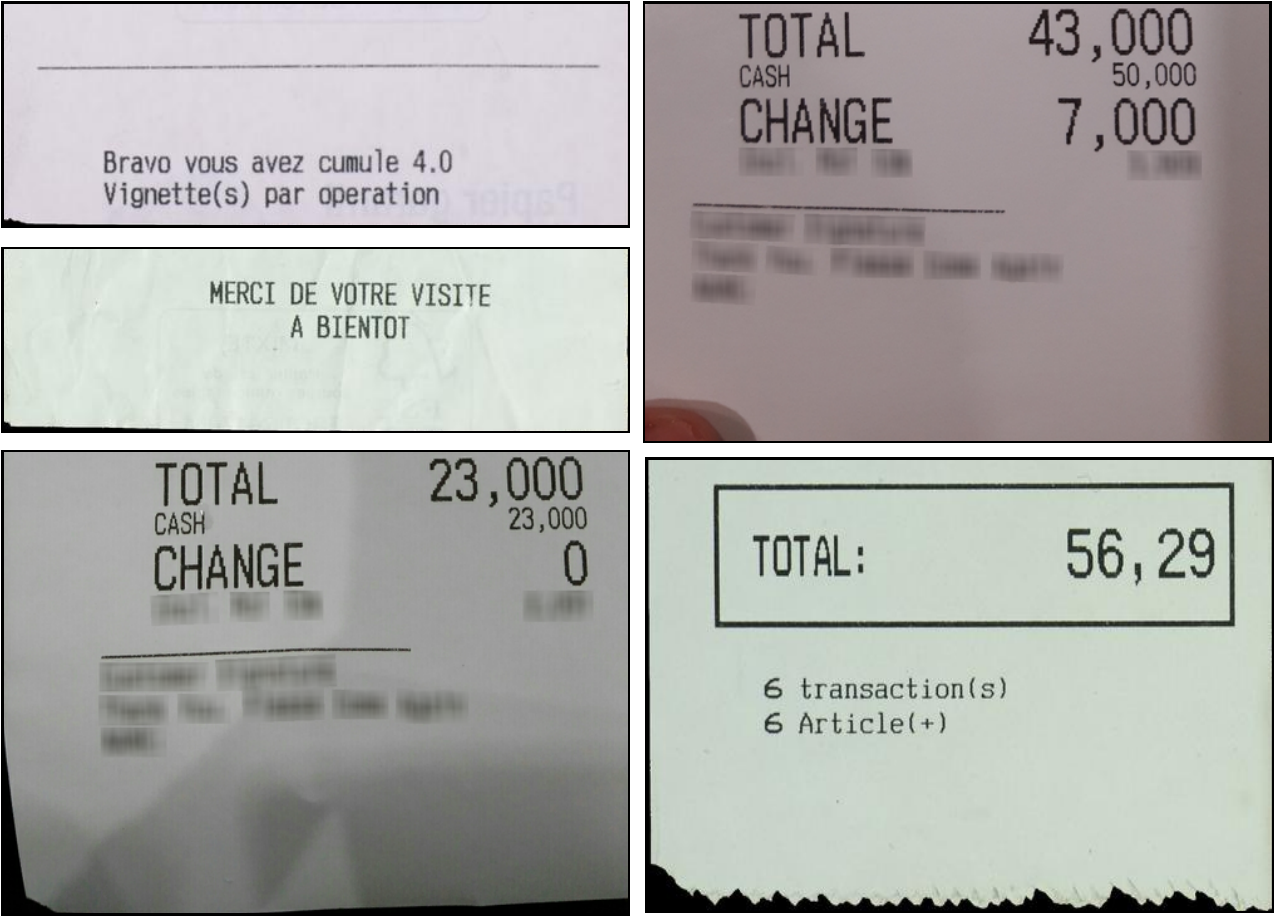}
        \vspace{0.5mm}
        \caption{Samples with 0 \textit{mean CER} that were pruned since they rank lowest with respect to Tesseract's CER.}
        \label{fig:prune_easy_imgs}
\end{figure}

We perform the data pruning experiments on the POS dataset with the Tesseract OCR engine. $n\%$ of the receipt images in the POS train set are pruned, where $n \in \{10, 20, 30, 40, 50\}$. The pruned dataset is used to train the system with single jitter, and the results are shown in Fig.  \ref{fig:pruning_results}. We can observe that the text recognition performance of Tesseract is on-par with the full dataset for $10\%$, $20\%$, and $30\%$ of data pruning. Further, pruning $30\%$ of the training dataset yields a $12.5\%$ reduction in OCR engine queries without any decrease in word-level accuracy. There is an implicit reduction in the number of queries since pruning document images also removes text strips that are never used to query the OCR engine. However, pruning more than $30\%$ of receipt images leads to a significant drop in test accuracy. Fig. \ref{fig:prune_easy_imgs} shows a few sample images from the pruned set. These images have less noise, less textual content, and legible fonts. Hence, Tesseract can accurately recognize the text in them without any preprocessing. 

\begin{figure}
        \centering
        \includegraphics[width=0.85\columnwidth]{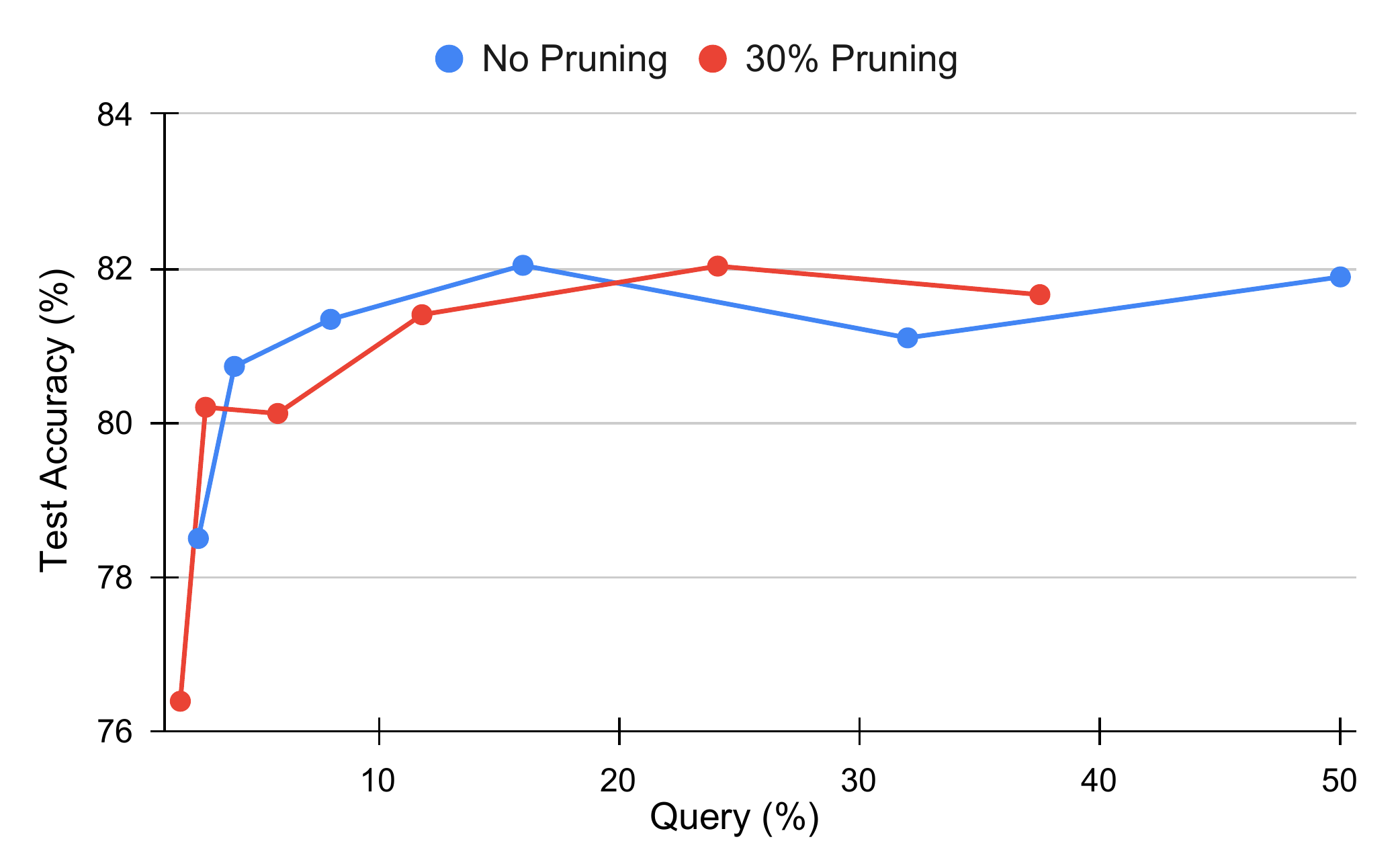}
        \vspace{0.5mm}
        \caption{Performance of system with original and pruned POS dataset using UniformCER sample selection for reducing queries to Tesseract.}
        \label{fig:pruning_comparison}
\end{figure}

We use the subset of data obtained by pruning $30\%$ of the images to train the preprocessor and reduce the number of OCR engine queries using UniformCER selection. We choose 2.5\%, 4\%, 8\%, 16\%, and 32\%  as the query budgets. It is important to note that performing query-efficient black-box approximation with a pruned dataset leads to a further reduction in OCR engine queries. For example, the 4\% query budget results in 2.8\%  of the total queries when the pruned dataset (30\% pruning) is used for training. Fig.  \ref{fig:pruning_comparison} depicts the results for different query budgets with and without data pruning. We can observe that the two curves are close, and the relationship between the curves changes for different (\%) query ranges. For instance, between 20\% and 40\% query budgets, the preprocessor trained with the pruned dataset performs better than the original dataset. Further, the system's performance with the pruned dataset at a very low query budget is better than that of the original dataset (first point in both curves). However, the pruned dataset performs slightly worse than the original dataset between 3\% and 20\% query budgets. This inconsistency in the results indicates that the receipt images can be selected with a better criterion so that the total OCR queries can be reduced by combining dataset pruning and query-efficient black-box approximation without compromising text recognition performance.

\section{Conclusion and Future Work}
In this paper, we propose a sample selection scheme that drastically reduces the number of queries to an OCR engine for efficiently training an OCR preprocessor using differentiable bypass. We demonstrate that using the selection algorithms with very low query budgets can significantly boost the text recognition performance for both open-source and commercial OCR engines. We also show that increasing the query budget leads to significantly larger training time with a small improvement in OCR performance. For future work, we want to explore techniques for querying the OCR engine with document images instead of individual text strips to reduce the total OCR engine queries. 

\section{Acknowledgement}
This research work was supported by Intuit AI Research and the Natural Sciences and Engineering Research Council of Canada (NSERC). 

\appendix

\bibliographystyle{elsarticle-num-names} 
\bibliography{references}





\newpage

\textbf{Ganesh Tata} is a Master’s student in the Department of Computing Science at the University of Alberta. His research interests include computer vision, machine learning, deep learning, and data subset selection.

\textbf{Katyani Singh} is a graduate student at the Department of Computing Science, University of Alberta. Her research interests include computer vision, image processing and deep learning. 

\textbf{Eric Van Oeveren} received his PhD in physics in 2018 from the University of Wisconsin-Milwaukee. He is now a Senior Data Scientist at Intuit, and works on machine learning models involving document intelligence.

\textbf{Nilanjan Ray} received his Ph.D. degree in electrical engineering from the University of Virginia, USA, in 2003. He is currently a Professor of computing science with the University of Alberta, Canada. His research interests include computer vision and medical image analysis. 
\end{document}